\documentclass{article} 
\usepackage{times}
\usepackage[pdftex]{graphicx}
\usepackage{subfigure}
\usepackage[square, sort, comma, numbers]{natbib}
\usepackage{amsmath}
\usepackage{amsfonts}
\usepackage{algorithm}
\usepackage{algorithmic}
\usepackage[top=3cm,bottom=2.5cm,left=4cm,right=4cm]{geometry}

\newcommand{\lo}{o}
\newcommand{\lO}{O}

\newcommand{\Happ}{\widehat{H}}
\newcommand{\ud}{\mathrm{d}}
\newcommand{\argmin}[1]{\underset{#1}{\arg\min}\ }
\newcommand{\mspc}{\hspace*{7.pt}}
\newcommand{\wt}{{\widetilde{\omega}}}

\begin{document}

\begin{center}

\rule{\linewidth}{4pt}
\rule{0pt}{20pt}
\textbf{\Large Fast Approximation of Rotations and \\ Hessians matrices}

\vspace*{10pt}
\rule{\linewidth}{1pt}
\end{center}

\begin{tabular}{cc}
\textbf{Micha\"el Mathieu} & \textbf{Yann LeCun} \\
Courant Institute of Mathematical Sciences & Courant Institute of Mathematical Sciences \\
New York University & New York University \\
\texttt{mathieu@cs.nyu.edu} & \texttt{yann@cs.nyu.edu} \\
\end{tabular}

\vspace*{30pt}

\begin{abstract}
A new method to represent and approximate rotation matrices is
introduced.  The method represents approximations of a rotation matrix $Q$ 
with linearithmic complexity, \emph{i.e.}  with $\frac{1}{2}n\lg(n)$
rotations over pairs of coordinates, arranged in an FFT-like
fashion. The approximation is ``learned'' using gradient descent.
It allows to represent symmetric matrices $H$ as $QDQ^T$ where
$D$ is a diagonal matrix. It
can be used to approximate covariance matrix of Gaussian models in
order to speed up inference, or to estimate and track the inverse
Hessian of an objective function by relating changes in parameters to
changes in gradient along the trajectory followed by the optimization
procedure.  Experiments were conducted to approximate synthetic
matrices, covariance matrices of real data, and Hessian matrices of
objective functions involved in machine learning problems.
\end{abstract}

\section{Introduction}

Covariance matrices, Hessian matrices and, more generally speaking,
symmetric matrices, play a major role in machine learning. In density
models containing covariance matrices (\emph{e.g.} mixtures of
Gaussians), estimation and inference involves computing the inverse of
such matrices and computing products of such matrices with
vectors. The size of these matrices grow quadratically with the
dimension of the space, and some of the computations grow
cubically. This renders direct evaluations impractical in large
dimension, making approximations necessary. Depending on the problem,
different approaches have been proposed.

In Gaussian mixture models (GMM), multiple high-dimensional Gaussian
functions must be evaluated. In very large models with high dimension
and many mixture components, the computational cost can be
prohibitive.  Approximations are often used to reduce the
computational complexity, including diagonal approximations, low-rank
approximations, and shared covariance matrices between multiple
mixture components.

In Bayesian inference with Gaussian models, and in variational
inference using the Laplace approximation, one must compute
high-dimensional Gaussian integrals to marginalize over the latent
variables. Estimating and manipulating the covariance matrices and
their inverse can quickly become expensive computationally. 

In machine learning and statistical model estimation, objective
functions must be optimized with respect to high-dimensional
parameters. Exploiting the second-order properties of the objective
function to speed up learning (or to regularize the estimation) is
always a challenge when the dimension is large, particularly when the
objective function is non quadratic, or even non convex (as is the
case with deep learning models). 

In optimization, quasi-Newton methods such as BFGS have been proposed
to keep track the inverse Hessian as the optimization proceeds. While
BFGS has quadratic complexity per iteration, Limited-Storage BFGS
(LBFGS) uses a factorized form of the inverse Hessian approximation
that reduces the complexity to linear times an adjustable
factor~\cite{nocedal-1980}. Quasi-Newton methods have been
experimented with extensively to speed up optimization in machine
learning, including LBFGS ~\cite{becker-lecun-89,lecun-98b,
  schraudolph-2007,bordes-bottou-gallinari-2009,ngiam-2011}, although
the inherently batch nature of LBFGS has limited its applicability to
large-scale learning~\cite{dean-nips-2012}. Some authors have
addressed the issue of approximating the Hessian for ML systems with
complex structures, such as deep learning architectures. This
involves back-propagating curvatures through the computational graph
of the function~\cite{becker-lecun-89,lecun-98b, chapelle-erhan-2011,
  martens-2012}, which is particularly easy when computing diagonal
approximations or Hessians of recurrent neural nets. Others have
attempted to identify a few dominant eigen-directions in which the
curvatures are larger than the others, which can be seen as a low-rank
Hessian approximation~\cite{lecun-simard-pearlmutter-93}

More recently, decompositions of covariance matrices using products of
rotations on pairs of coordinates have been explored
in~\cite{bouman-2011}.  The method greedily selects the best pairwise
rotations to approximate the diagonalizing basis of the matrix, and
then finds the best corresponding eigenvalues.

\section{Linearithmic Symmetric Matrix Approximation}

A symmetric matrix $H$ of size $n\times n$ can be factorized
(diagonalized) as $H=QDQ^T$ where $Q$ is an orthogonal (rotation)
matrix, and $D$ is a diagonal matrix. In general, multiplying a vector
by $Q$ requires $n^2$ operations, and the full set of orthogonal
matrices has $n(n-1)/2$ free parameters. The main idea of this paper
is very simple: parameterize the rotation matrix $Q$ as a product of
$n\lg(n)/2$ elementary rotations within 2D planes spanned by pairs of
coordinates (sometimes called Givens rotations). This makes the
compuational complexity of the product of $H$ by a vector to
linearithmic instead of quadratic. It also makes the computation of
the inverse trivial $H^{-1} = Q^T D^{-1} Q$. The second idea is to
compute the best such approximation of a matrix by least-square
optimization with stochastic gradient descent.

The question is how good of an approximation to real-world covariance
and Hessian matrices can we get by restricting ourselves to these
linearithmic rotations instead of the full set of rotations.

The main intuition that the approximation may be valid comes from a
conjecture in random matrix theory according to which the distribution
of random matrices obtained randomly drawing $n\lg(n)/2$ successive
Givens rotations is dense in the space of rotation
matrices~\cite{benarous-2013}. Proven methods for generating
orthogonal matrices with a uniform distribution require
$O(\log(p)p^2)$ Givens rotation, with an arrangement based on the
butterfly operators in the Fast Fourier Transform~\cite{genz-1998}.

Indeed, to guarantee that pairs of coordinates are shuffled in a
systematic way, we chose to arrange the $n\lg(n)/2$ Givens rotations
in the same way as the butterfly operators in the Fast Fourier
Transform. To simplify the discussion, we assume from now on that $n$
is a power of $2$, unless specified otherwise. We decompose
\begin{equation}
Q \approx Q_1Q_2\dots Q_{\lg(n)}
\end{equation}
where $Q_i$ is a sparse rotation formed by $n/2$ independent pairwise
rotations, between pairs of coordinates
$$\big(2pk+j,\qquad p(2k+1)+j\big)$$
where $p=\frac{n}{2^i}$, $k\in0..2^{i-1}$ and $j\in1..p$.
For instance, with $n=8$, we obtain the following matrices :
\begin{gather*}
Q_1 =
\left(
\begin{smallmatrix}
  \mspc{}c_1 & & & & -s_1 & & &\\
  & \mspc{}c_2 & & & & -s_2 & &\\ 
  & & \mspc{}c_3 & & & & -s_3 &\\
  & & & \mspc{}c_4 & & & & -s_4\\
  \mspc{}s_1 & & & & \mspc{}c_1 & & &\\
  & \mspc{}s_2 & & & & \mspc{}c_2 & &\\
  & & \mspc{}s_3 & & & & \mspc{}c_3 &\\
  & & & \mspc{}s_4 & & & & \mspc{}c_4\\
\end{smallmatrix}
\right)
\end{gather*}
\begin{gather*}
Q_2 =
\left(
\begin{smallmatrix}
  \mspc{}c_5 & & -s_5 & & & & &\\
  & \mspc{}c_6 & & -s_6 & & & &\\ 
  \mspc{}s_5 & & \mspc{}c_5 & & & & &\\
  & \mspc{}s_6 & & \mspc{}c_6 & & & &\\ 
  & & & & \mspc{}c_7 & & -s_7 &\\
  & & & & & \mspc{}c_8 & & -s_8\\
  & & & & \mspc{}s_7 & & \mspc{}c_7 &\\
  & & & & & \mspc{}s_8 & & \mspc{}c_8\\
\end{smallmatrix}
\right)\\
Q_3 =
\left(
\begin{smallmatrix}
  \mspc{}c_9 & -s_9 & & & & & &\\
  \mspc{}s_9 & \mspc{}c_9 & & & & & &\\
  & & \mspc{}c_{10} & -s_{10} & & & &\\
  & & \mspc{}s_{10} & \mspc{}c_{10} & & & &\\
  & & & & \mspc{}c_{11} & -s_{11} & &\\
  & & & & \mspc{}s_{11} & \mspc{}c_{11} & &\\
  & & & & & & \mspc{}c_{12} & -s_{12}\\
  & & & & & & \mspc{}s_{12} & \mspc{}c_{12}\\
\end{smallmatrix}
\right)
\end{gather*}
where $c_i = \cos(\theta_i)$ and $s_i=\sin(\theta_i)$.

There are $n\lg(n)/2$ pairwise rotations, grouped into $\lg(n)$
matrices, such that each matrice represents $n/2$ independent
rotations. It is the minimal number of matrices that keeps the
pairwise rotations structure, and this particular arrangement
has the property that there can be an interaction between each
pair of coordinates of the input.

The final decomposition of the symmetric matrix $H$
is therefore a sequence of $2\lg(n)+1$ sparse matrices, namely
\begin{equation}
\label{hdecomp}
H\approx Q_1Q_2\dots Q_{\lg(n)}DQ_{\lg(n)}^T\dots Q_2^TQ_1^T
\end{equation}

Since all the matrices $Q_i$ are rotations, we have $Q_i^{-1}=Q_i^{T}$ and
thus the decomposition of $H^{-1}$ is the obtained by simply replacing
the elements of $D$ by their inverse.

It is important to notice that the problem becomes easier if $H$ has
eigenspaces of high dimension. If there are groups of eivenvalues roughly
equal, the rotations between vectors belonging to the corresponding eigenspaces
become irrelevant. This is similar to low rank approximation, where all small
eigenvalues are grouped into the same eigenspace, but it can work for
eigenspaces with non-zero eigenvalues.

\section{Approximating Hessian matrices}

\subsection{Methodology}

Let $(\theta_1, \dots, \theta_{n\lg(n)/2})$ be the rotation parameters of the
matrices $Q_1, \dots, Q_{\lg(n)}$ and $(\sigma_1, \dots, \sigma_n)$ the diagonal
elements of $D$. Let us name $\omega$ the
whole set of $n(\lg(n)/2+1)$ parameters, so
$$\omega=(\theta_1, \dots, \theta_{n\lg(n)/2}, \sigma_1, \dots, \sigma_n)$$
When we need the explicit parametrization, we denote $Q_{\omega}$ the rotation
matrix $Q_1\dots Q_{\lg(n)}$ parametrized by $(\theta_j)$
and $D_{\omega}$ the diagonal matrix $\mathrm{diag}(\sigma_1,\dots,\sigma_n)$.
Finally, let $\mathcal{Q} = \{Q_{\omega}|\omega\in\mathbb{R}^{n(\lg(n)+1)}\}$ be
the set of all matrices $Q_{\omega}$.

In order to find the best set of parameters $\omega$, we first notice
that the sequence of matrices in the decomposition (\ref{hdecomp}) can be seen
as a machine learning problem.
Although the problem is unfortunately not convex, it can still be minimized
by using standard gradient descent techniques, such that SGD or minibatch
gradient descent.

Let $(x^{(j)})_{j=1..m}$ be a set of $n$-dimensional vectors. We train our model
to predict the output $Hx^{(j)}$ when given the input $x^{(j)}$. We use a least
square loss, so the function we minimize is
\begin{eqnarray}
\mathcal{L}(\omega) &=& \sum_{j=1}^m ||Q_{\omega}D_{\omega}Q_{\omega}^Tx^{(j)}-y^{(j)}||_2^2\\
&=&\sum_{j=1}^m L(\omega, x^{(j)}, y^{(j)})
\end{eqnarray}
where $y^{(j)} = Hx^{(j)}$ and
\begin{equation}
L(\omega, x^{(j)}, y^{(j)}) = ||Q_{\omega}D_{\omega}Q_{\omega}^Tx^{(j)}-y^{(j)}||_2^2
\end{equation}

However, the parameterization through the angles makes the problem complicated
and hard to optimize. Therefore, during one gradient step, we relax the
parametrization and allow each matrix $(Q_i)$ to have $2n$ independent
parameters, and we reproject the matrix on the set of the rotation matrices
after the parameter update. The problem becomes learning a
multilayer linear neural network, with sparse connections and shared weights. \\
Formally, we use an extended set of parameters, $\wt$,
which has one parameter per non-zero element of each matrix $Q_i$.
The new set of matrices parametrized by $\wt$, which
we denote $\widetilde{\mathcal{Q}}$, contains
matrices which are not rotation matrices. However, because of the Givens
structure inside the matrices, it is trivial to project a matrix from
$\widetilde{\mathcal{Q}}$ on the set $\mathcal{Q}$, by simply projecting every
Givens on the set of the $2\times2$ rotations.
The projection $proj$ of a Givens is
\begin{equation}
\label{eqn:projgivens}
proj\begin{pmatrix}
a & b \\
c & d \\
\end{pmatrix}
=\frac{1}{\eta}\begin{pmatrix}
a+d \quad & b-c\\
c-b \quad & a+d\\
\end{pmatrix}
\end{equation}
where $\eta=\sqrt{(a+d)^2+(b-c)^2}$.

We summarize the learning procedure in Algorithm~\ref{alg:learnmatrix}, using SGD.
Computing the gradient of $L(\wt, x^{(j)}, y^{(j)})$ is performed by a gradient
backpropagation. Note that we don't need the matrix $H$,
but only a way to compute the products
$y^{(i)}=Hx^{(i)}$.

The hyperparameter $\alpha$, which doesn't have to be constant,
is the learning rate. There is no reason that the same learning rate should
be applied to the part of $\wt$ in $Q_{\wt}$ and
to the part in $D_{\wt}$. Therefore, $\alpha$ can be a diagonal
matrix instead of a scalar. We have observed that the learning
procedure converges faster when the
part in $D_{\wt}$ has a smaller learning rate.

\begin{algorithm}[ht]
  \caption{Hessian matrix learning}
  \label{alg:learnmatrix}
\begin{algorithmic}
  \STATE{\bfseries Input:} set $(x^{(j)}, y^{(j)})$ for $j=1..m$
  \WHILE{not converged}
  \STATE{Randomly draw $j\in 1..m$}
  \STATE{Compute gradient $g_{\wt,j} = \frac{\partial L(\wt,x^{(j)}, y^{(j)})}{\partial \wt}$}
  \STATE{Update $\wt\leftarrow \wt - \alpha g_{\wt,j}$}
  \STATE{Normalize all the Givens to project $Q_{\wt}$ on $\mathcal{Q}$}
  \ENDWHILE
\end{algorithmic}
\end{algorithm}

There are several ways to obtain the input vectors $x^{(j)}$, depending on the
intended use of the approximated matrix. If we only need to obtain a fast and
compact representation, we can draw random vectors. We found that vectors
uniformly sampled on the unit sphere or in a hypercube centered on zero
perform well. In certain cases,
we may need to evaluate products $Hx$ for $x$ in a certain region of
$\mathbb{R}^n$. In this case, using vectors in this specific area can provide
better results, 
since more of the expressivity of
our decomposition will be used on this specific region.
The set of vectors $(x^{(j)}, y^{(j)})$ can also appear naturally, as we can
see in the next part when we approximate the Hessian of another optimization
problem.

\subsection{Experiments}

In order to confirm our hypothesis, we ran our algorithm on synthetic and natural
Hessian matrices. Synthetic matrices are obtained by randomly drawing rotations and
eigenvalues, while the natural matrix we tried is the Hessian of the MNIST dataset.

To evaluate the quality of the approximation, we measure the average angle
between the $QDQ^Tx$ and $Hx$ for a set of random vectors $x$.

\begin{figure}[ht]
\vskip 0.2in
\begin{center}
\centerline{\includegraphics[width=\columnwidth]{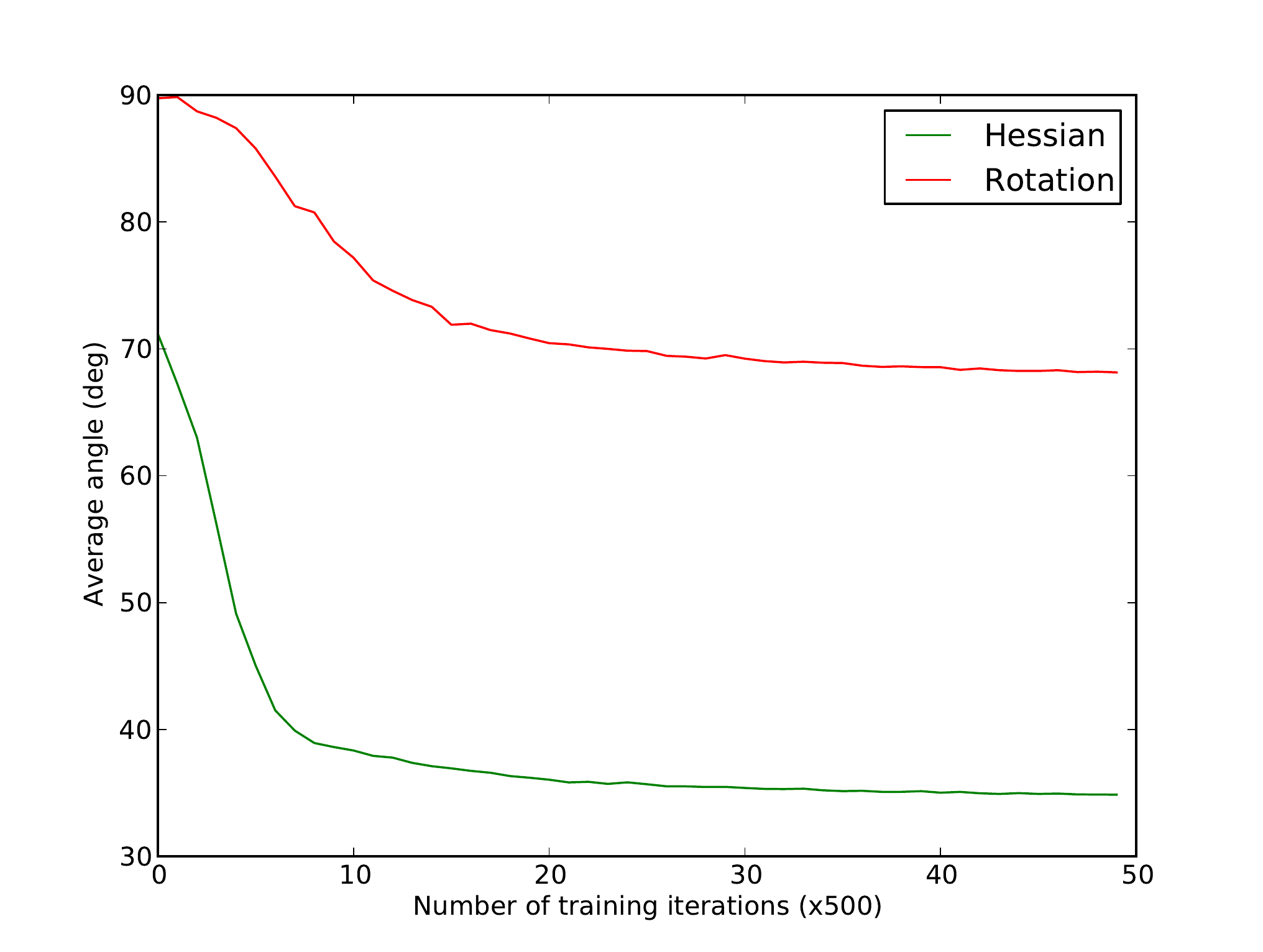}}
\caption{Average error angle versus epochs for matrix learning. In
  green, learning $H$, with the number of larger eigenvalues
  $n_{\mu}=5$. In red, learning a rotation matrix $R$.}
\label{fig:approxh}
\end{center}
\vskip -0.2in
\end{figure}

The synthetic matrices $H$ are generated as $R\Lambda R^T$ where $R$
is a random rotation matrix, uniformly drawn on the set of rotation
matrices, and $\Lambda = \mathrm{diag}(\lambda_i)$ is a diagonal
matrix, with $\lambda_i=|\mu_i|$ where $\mu_i$ is sampled from a
zero-mean and $0.1$ Gaussian distribution, except a small number
$n_{\mu}$ of randomly selected $\mu_i$ that are drawn from a
Gaussian of mean $1$ and variance $0.4$. \\ The results are displayed
on Figure~\ref{fig:approxh}. The average angle goes down to about $35$
degrees in a space of dimension $64$, with the number of large
eigenvalues $n_{\mu}=5$.

For sanity check purposes, we also try directly learning a random rotation matrix
$R$ using a single $Q$ matrix. The loss function is then
\begin{eqnarray}
  \mathcal{L}'(w) = \sum_j||Qx^{(j)} - Rx^{(j)}||_2^2
\end{eqnarray}
The results are also shown on Figure~\ref{fig:approxh}. It doesn't perform as well
as learning a Hessian, which was expected since we don't have eigenspaces of
high dimension, but it still reaches $68$ degrees in dimension $64$. \\
We performed another sanity check : if the matrix $R$ is actually the form
of a matrix $Q_{\omega}$, then we learn it correctly (loss function $\mathcal{L}'$
reaching zero), and we also learn $H=R^T\Lambda R$ with
loss function $\mathcal{L}$ going quickly to zero.

\begin{figure}[ht]
\vskip 0.2in
\begin{center}
\centerline{\includegraphics[width=\columnwidth]{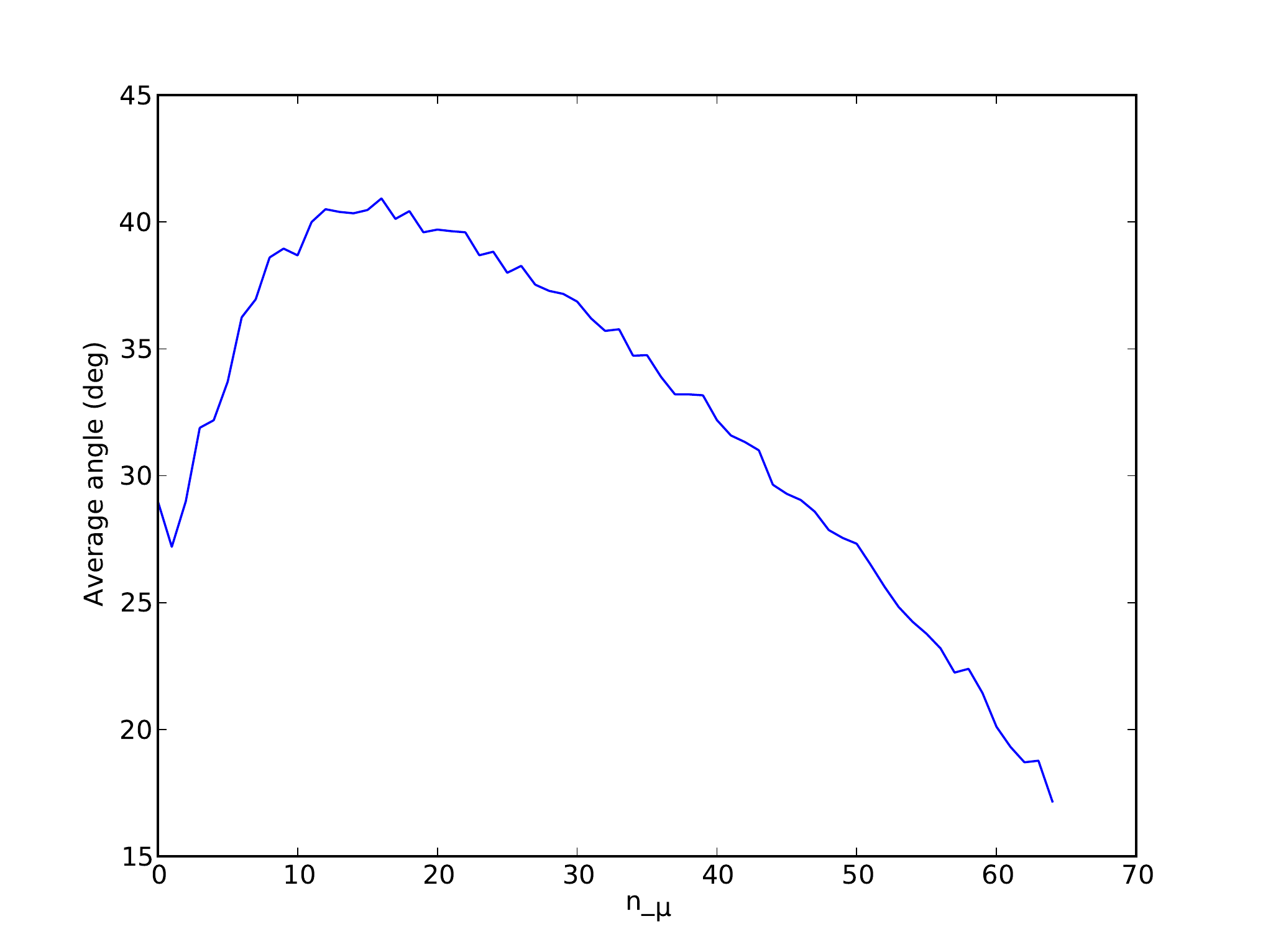}}
\caption{Average error angle after convergence versus the number of
  dominant eigenvalues $n_{\mu}$.}
\label{fig:dvalues}
\end{center}
\vskip -0.2in
\end{figure}

It appears that the number of larger eigenvalues $n_{\mu}$ has an
influence on the quality of the approximation : the average error
angle after convergence is lower when
$n_{\mu}$ is close to $0$ or to the size of the space (about $20$ degrees in the
latter case). On the other
hand, when the eigenvalues get split into two groups of about the same
size, the average error angle can is about $41$ degrees. The average
angle as a function of $n_{\mu}$ is shown on Figure~\ref{fig:dvalues}.

\begin{figure}[ht!]
\vskip 0.2in
\begin{center}
\centerline{\includegraphics[width=0.5\linewidth]{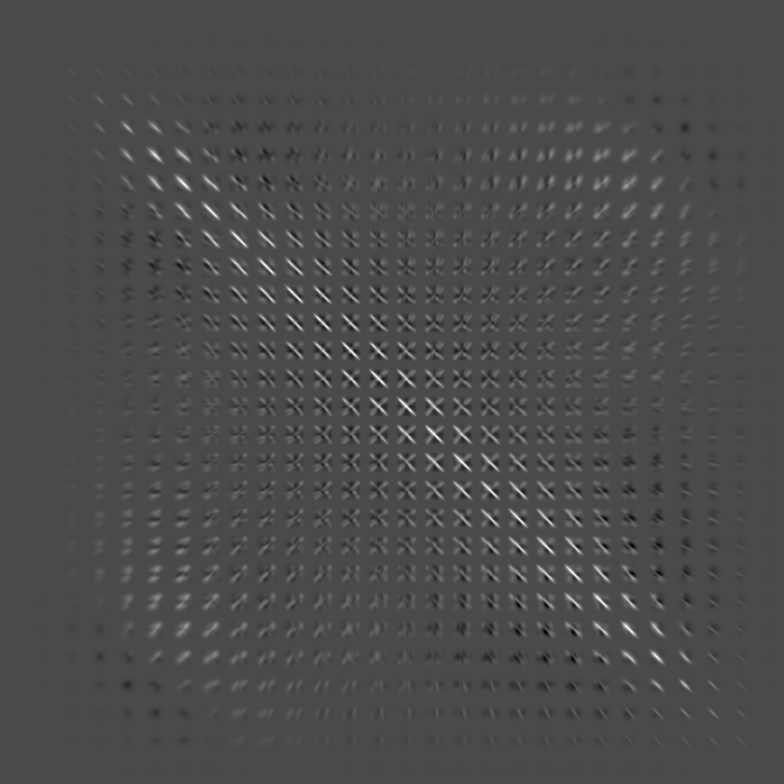}
\includegraphics[width=0.5\linewidth]{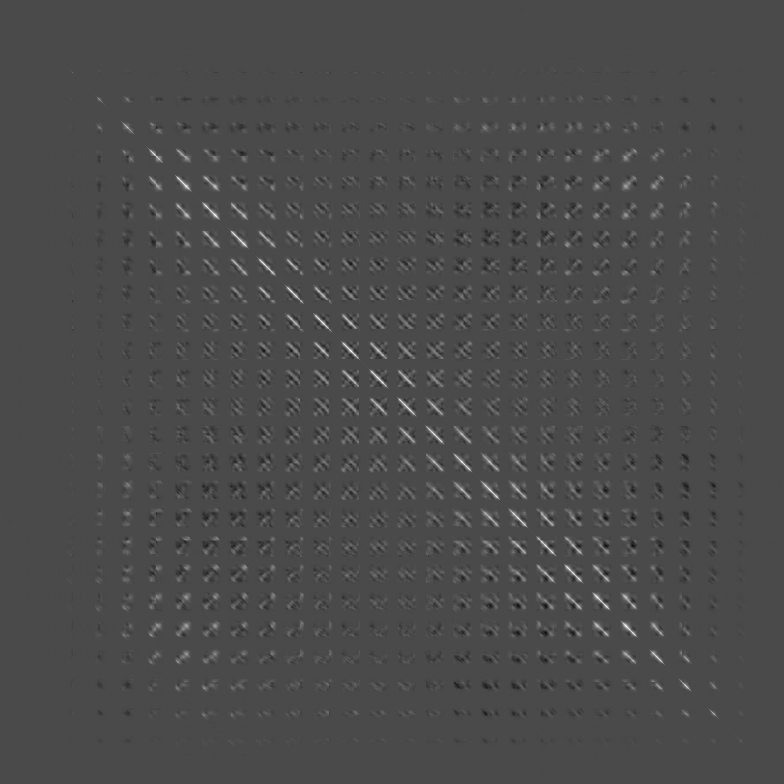}}
\caption{True (left) and approximated (right) covariance matrix of the MNIST dataset.}
\label{fig:mnist-cov}
\end{center}
\vskip -0.2in
\end{figure}

We tested our algorithm on the covariance of the MNIST dataset. It reached an angle
of $38$ degrees. Figure~\ref{fig:mnist-cov} shows real and approximated covariance
matrices.

\section{Hessian matrices of loss functions}

In this section, we denote $\Happ=Q^TDQ$ (or $\Happ_{\wt}=Q^T_{\wt}D_{\wt}Q_{\wt}$) the approximated Hessian.

\subsection{Batch gradient descent}

Although learning an approximation of a fixed Hessian matrix has uses,
our method can be used in an online fashion to track slowly-varying
Hessian matrices over the course of an optimization procedure. 

In numerous cases, we have an optimization problem involving too many variables
to explicitly compute the Hessian, preventing us from using direct second order
methods. 
Our Hessian representation can be used to store and evaluate
the Hessian matrix. Thanks to the learning procedure involved, if the Hessian
of the loss function changes, our estimation will be able to adapt.

Let us consider an optimization problem where we aim to minimize a loss function
$\ell(u)$ where $u\in\mathbb{R}^n$ is the set of parameters.
Therefore, we are looking for
\begin{equation}
u^*=\argmin{u\in\mathbb{R}^n}\ell(u)
\end{equation}
Or, in case of a non convex problem, we are looking for a $u^\dagger$ that gives
a ``low'' value for $\ell$ (``low'' being problem dependent). \\
In addition, let us suppose we are in a case where it is possible to compute
the gradient $\nabla_u\ell$ of $\ell$ at every point $u$. It is the case, for
instance, for neural networks, using the backpropagation algorithm, or whenever
we have an efficient way to compute the gradient.

It is well known 
that even if the problem is quadratic,
and therefore the Hessian matrix $H$ doesn't depend on the point $u$,
a gradient descent algorithm can perform poorly when $H$ has
eigenvalues with different magnitude. This situation becomes even
worse when the loss function is not quadratic in the parameters and
the Hessian is not constant (we use notation $H_u$ for the Hessian
matrix at point $u$). However, if $H_u$ is known, we can replace the
gradient step $\nabla_u\ell$ by $H_u^{-1}\nabla_u\ell$. This modifies
the descent direction to take the curvature into account and allows
quadratic convergence if the function is locally quadratic. Since
$H_u$ is too large to compute or store, we can rely on approximations,
such as in LBFGS or Truncated Conjugate Gradient.

Our method provides another way to track an approximation of $H_u$.
By definition, we have
\begin{equation}
H_{u+\ud{}u}\ud u = \nabla_{u+\ud{}u}\ell - \nabla_u\ell + \lo(||\ud u||)
\end{equation}
In the context of minimization, let us suppose we have taken a step from
point $u_{t-1}$ and arrived at the current point $u_t$. Then we know that
\begin{equation}
\label{eqn:trainhessian}
H_{u_t}(u_t-u_{t-1}) \approx \nabla_{u_t}\ell - \nabla_{u_{t-1}}\ell
\end{equation}
We can use this relation as a training sample for the matrix $H$.

Thus, we start with an approximated matrix $H_{u_0}$ set to identity,
and we begin a regular gradient descent. At each step, we obtain a training
point as defined in Equation~\ref{eqn:trainhessian}, and we perform one step
in learning $H_{u_1}$ (following the gradient of the loss function $L$).
Then we perform a second step in the minimization of $\ell$, folloing vector
$H_{u_1}^{-1}\cdot\nabla_{u_1}\ell$, and so on. Note that our representation
of $H_{u_t}$ allows us to compute the inverse easily. The procedure is
summarized in Algorithm~\ref{alg:gradientdescenthessian}.

\begin{algorithm}[ht]
  \caption{Optimization with linearithmic Hessian}
  \label{alg:gradientdescenthessian}
  \begin{algorithmic}
    \STATE{\bfseries Parameters:} Learning rates $\alpha$ and $\beta$
    \STATE{Initialize $\wt$ such that $D_{\wt}=I$
      and $Q_{\wt}\in \mathcal{Q}$}
    \STATE{Initialize random $u_0$}
    \STATE{Set $t=0$}
    \WHILE{not converged}
    \STATE{Compute $\nabla_{u_t}\ell$}
    \IF{$t\neq 0$}
    \STATE{Set $\delta u = u_t - u_{t-1}$}
    \STATE{Set $\delta g = \nabla_t \ell - \nabla_{t-1} \ell$}
    \STATE{Update $\wt\leftarrow\wt-\alpha\frac{\partial L(\wt, \delta u, \delta g)}{\partial\wt}$}
    \STATE{Project $Q_{\wt}$ on $\mathcal{Q}$ as in Equation~\ref{eqn:projgivens}}
    \ENDIF
    \STATE{Set $u_{t+1} = u_{t} - \beta \Happ_{\wt}\nabla_u\ell$}
    \STATE{Update $t\leftarrow t+1$}
    \ENDWHILE
  \end{algorithmic}
\end{algorithm}

A number of implementation details must be considered :
\begin{itemize}
  \item It appears that in natural Hessians, many of the eigenvalues are small
    or zero, especially when we approach a local minimum. Therefore, the inverse
    of $D$ can diverge. We solve this issue by imposing a minimal value $\epsilon$ on
    the elements of $D$ when computing their inverse.
  \item Similarly, eigenvalues can be negative. It is a problem in the minimization
    since it makes second order methods diverge if not taken into account. There
    are several ways to deal with such eigenvalues. We chose to set negative
    elements of $D$ to $\epsilon$ at the same time as the normalization of
    $Q_{\wt}$.    
  \item The norm of $\delta u$ (as defined in
    Algorithm~\ref{alg:gradientdescenthessian}) can vary from one step to the other.
    However, since the Hessian approximation is linear, scaling $\delta u$ and
    $\delta g$ by the same factor is equivalent to scaling the learning rate
    $\alpha$ by the same factor. To have a better control on the learning rate,
    we rescale both $\delta u$ and $\delta g$ by a factor $1/||\delta u||$.
\end{itemize}

Although it performs poorly, due to the small eigenvalues, it is also possible
to directly learn the inverse Hessian matrix using the relation
\begin{equation}
\delta u \approx H_u^{-1} \delta g
\end{equation}

\subsection{Stochastic and minibatch gradient descent}

For most problems, stochastic gradient descent (SGD) performs better than
batch descent. Here, we will consider the intermediate case, minibatch descent,
where each step uses only a part of the dataset. It covers both plain SGD and
batch methods by setting the size of the minibatch to $1$ or to the size of the
dataset.

In minibatch setting, the loss function becomes $\ell_A(u)$, where $A$ is a subset
of the dataset, changing at each iteration. We have $\ell_A(u)=\ell(u)$ if $A$ is
the whole dataset. Because of this difference, the previous analysis doesn't
hold anymore, since the Hessian of $\ell_A$ is not going to be the same as
the Hessian of $\ell_{A'}$.

However, we will see that if we use the Hessian learning in this setting,
it still perform something very similar to minibatch gradient,
on the Hessian approximation. When learning the approximation,
instead of training pairs $(\delta u, \delta g)$, we now obtain
pairs $(\delta u, \delta g_A)$, where $\delta u = u_t - u_{t-1}$ and
$\delta g_A = \nabla_{u_t} \ell_A - \nabla_{u_{t-1}} \ell_A$. Let $\mathcal{A}$
be the whole dataset and $(A_i)_{i=1..p}$ be minibatches (so $A_i\subset\mathcal{A}$).
Since the gradient operator is linear, if $\mathcal{A} = \uplus_{i=1}^p A_i$, then
\begin{equation}
\nabla_u \ell = \frac{1}{|\mathcal{A}|}\sum_{i=1}^p \nabla_u \ell_{A_i}
\end{equation}
This simply states that if we sum gradients, taken at the same point,
from a set of minibatches such that the dataset is their disjoint union,
then we obtain the batch gradient. We have scaled the
batch gradient by the size of the dataset for clarity purposes in the
following part.

We will show that the same property holds for the loss function $L$. The
loss function for the Hessian approximation corresponding to minibatch
$A_i$ is $L(\wt, \delta u, \delta g_{A_i})$. Its gradient is
\begin{equation}
\frac{\partial L(\wt, \delta u, \delta g_{A_i})}{\partial \wt}
     = 2\frac{\partial \Happ_\wt}{\partial \wt}\delta u (\Happ_\wt \delta u - \delta g_{A_i})
\end{equation}
So we get
\begin{eqnarray}
  &&\frac{1}{|\mathcal{A}|}\sum_{i=1}^p\frac{\partial L(\wt, \delta u, \delta g_{A_i})}{\partial \wt}\\
  & = & 2\frac{\partial \Happ_\wt}{\partial \wt}\delta u (\Happ_\wt \delta u - \sum_{i=1}^p\delta g_{A_i})\\
  & = & 2\frac{\partial \Happ_\wt}{\partial \wt}\delta u (\Happ_\wt \delta u - \delta g_{\mathcal{A}})\\
  & = & \frac{\partial L(\wt, \delta u, \delta g_{\mathcal{A}})}{\partial \wt}
\end{eqnarray}

Another issue with minibatch descent lies in the difficulty to reuse the gradient
computed at step $t-1$ to compute $\delta g$ at step $t$. Indeed,
$\delta g_{A_i} = \nabla_{u_t} \ell_{A_i} - \nabla_{u_{t-1}} \ell_{A_i}$. In the
batch method, we had already computed $\nabla_{u_{t-1}} \ell_A$ at the previous
step, and we could reuse it. With the minibatch approach, the previous step
used $\nabla_{u_{t-1}} \ell_{A_{i-1}}$, which is not computed on the same minibatch.\\
The easiest way to overcome this issue is to recompute $\nabla_{u_{t-1}} \ell_{A_{i}}$
at step $t$, but it implies doubling the number of gradients computed. Although
we don't have a fully satisfying answer to this problem, a possibility to reduce
the number of gradients computed would be to use the same minibatch $A_i$ a few times
consecutively before changing to $A_{i+1}$.

\subsection{Performance}

The approximated Hessians are sparse matrices with $2n$ coefficients each.
They can be efficiently implemented using standard sparse linear algebra,
such as \texttt{spblas}. The total number of operations for evaluating
an approximate Hessian is $\lO(n\lg(n))$. If we neglect the fact that sparse
matrices involve more operations, one evaluation coses $4n\lg(n)+n$
\emph{MulAdd} operations.
Computing the gradient of $\Happ$ has the same cost, thanks to the backpropagation
algorithm. Accumulating the gradients cost another $4n\lg(n)+n$. So tracking the
Hessian matrix using our method adds about $12n\lg(n)+3n$ operations at each
iteration and, when using minibatches, possibly another gradient evaluation.

In terms of memory, there are $2n\lg(n)+n$ parameters to store, and another
$2n\lg(n)+n$ is required to store the gradient of $\Happ$ during learning.

\section{Applications and future work}

The linearithmic structure makes the Hessian factorization scalable to larger
models. In particular, large inference problems invloving Gaussian functions
are good candidates for our method. Evaluating a product $x^TR^TDRx$ is
even faster, since it can be rewritten $(Rx)^TD(Rx)$, and therefore be performed in
$2n\lg(n)+2n$ operations. This is particularly adapted to inference, where it is
possible to spend some time to learn the approximate the matrix once and for all.
Gaussian mixture models and Bayesian inference with Gaussian models could
profit from this approximation.

Although some work has still to be done, speeding up optimization by tracking
the approximated Hessian could help optimizing function. In particular, deep
neural network could profit from this method, since their loss function tend
to be non isotropic, so they could profit from fast approximated second order
methods.

\newpage
\bibliographystyle{alpha}
\bibliography{rotations}

\end{document}